\documentclass{article}
\usepackage{graphicx} 
\usepackage{booktabs}
\usepackage{longtable}
\usepackage{url}
\usepackage{array}
\newcolumntype{C}[1]{>{\centering\arraybackslash}m{#1}}
\usepackage{pgfplots}
\usepackage{enumitem}

\title{\includegraphics[width=0.08\textwidth]{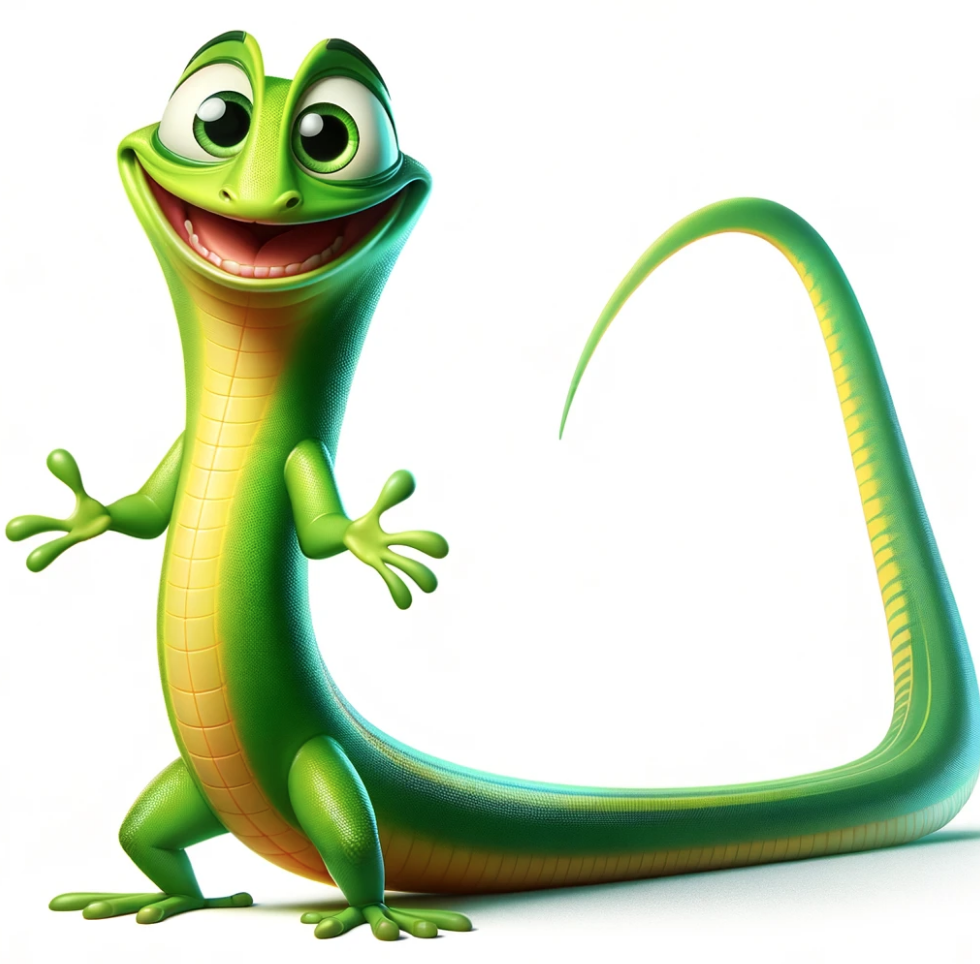} Lissard: Long and Simple Sequential Reasoning Datasets}
\author{Mirelle Bueno,
Roberto Lotufo, and Rodrigo Nogueira\\
\small FEEC, UNICAMP, Brazil}

\date{}

\begin{document}

\maketitle
\begin{abstract}

Language models are now capable of solving tasks that require dealing with long sequences consisting of hundreds of thousands of tokens. However, they often fail on tasks that require repetitive use of simple rules, even on sequences that are much shorter than those seen during training. For example, state-of-the-art LLMs can find common items in two lists with up to 20 items but fail when lists have 80 items. 
In this paper, we introduce Lissard, a benchmark comprising seven tasks whose goal is to assess the ability of models to process and generate wide-range sequence lengths, requiring repetitive procedural execution. Our evaluation of open-source (Mistral-7B and Mixtral-8x7B) and proprietary models (GPT-3.5 and GPT-4) show a consistent decline in performance across all models as the complexity of the sequence increases. The datasets and code are available at \url{https://github.com/unicamp-dl/Lissard}


\end{abstract}

\section{Introduction}
\begin{figure}[h] 
    \begin{center}
        \includegraphics[width=1\textwidth]{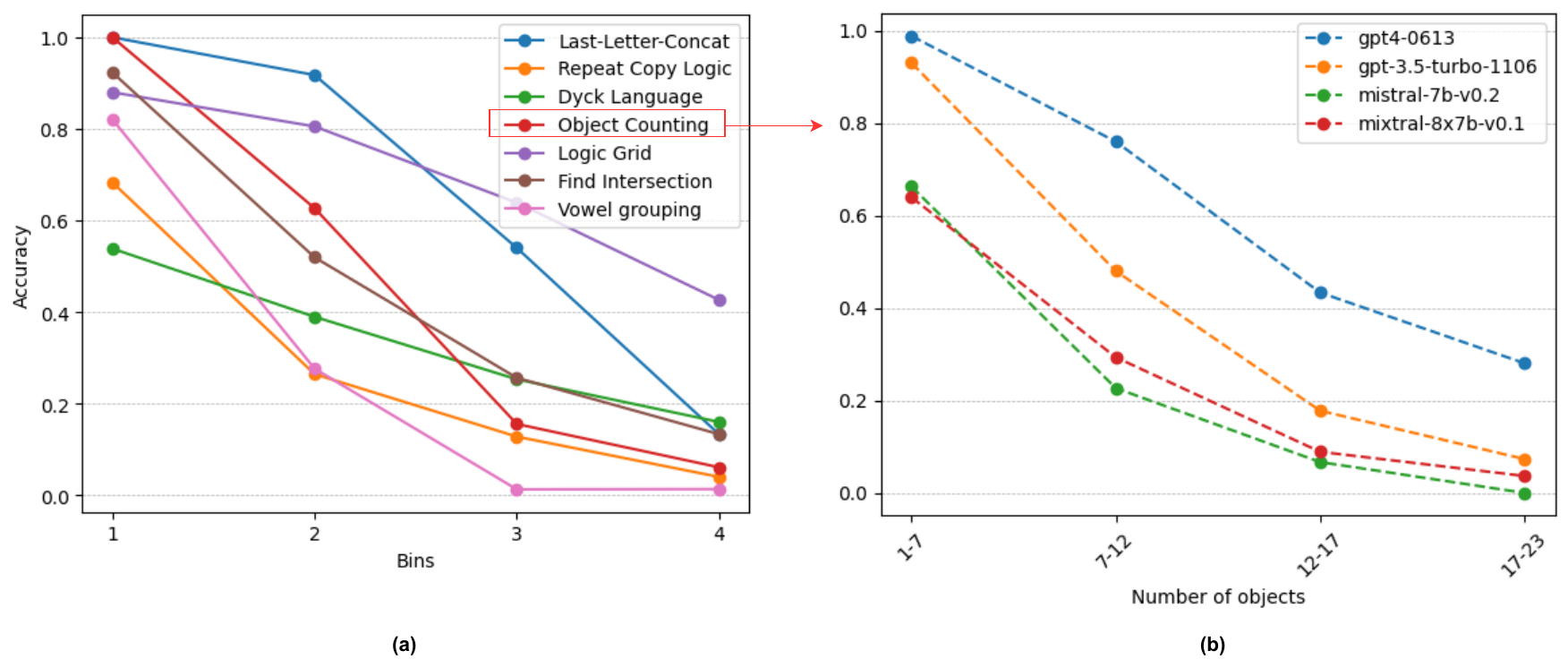}
    \end{center}
    \vspace{-0.5cm}
    \caption{ (a) Performance of GPT-4 on the Lissard benchmark (see Table~\ref{tab:key_entity_bins} for the definition of the bins); (b) Comparative performance of models in the ``Object Counting'' task.}
    \label{fig:gpt_4_all_tasks}
\end{figure}
 
The efficacy of language models, particularly in reasoning tasks, is significantly impacted by longer text lengths than those seen in training \cite{li2023functional, anil2022exploring,lake2018generalization}. This phenomenon, referred to as ``Length Generalization'' or ``Length Extrapolation'' in the literature \cite{press2022train, zhao2023length}, is also common in models based on the Transformer architecture \cite{livska2018memorize,lewkowycz2022solving, delétang2023neural,zhou2023algorithms}. Notably, even Large Language Models (LLMs), known for their strong performance in a wide range of tasks and domains, are not immune to this problem \cite{anil2022exploring,chen2023extending}.

Recent research tried to address this challenge by modifications to the positional embeddings~\cite{press2022train, chi2022kerple,chi-etal-2023-dissecting, li2023functional,ke2021rethinking} or by using prompting strategies such as scratchpad~\cite{nye2021work} and chain-of-thought reasoning~\cite{wei2023chainofthought}. Nevertheless, there remains a lack of datasets specifically designed for the systematic evaluation of the problem.

While benchmarks such as SCROLLS \cite{shaham-etal-2022-scrolls} were designed to evaluate models in natural language tasks that involve long sequences, its effectiveness in monitoring model performance degradation within the context of length generalization may be limited by lack of explicit control of task complexity with respect to sequence length. For example, when using natural language texts there is no guarantee that answering a question about a longer text is harder than responding to one about a shorter text. This limitation highlights the need for benchmarks that can explicitly manipulate and test the impact of sequence length on model performance. In benchmarks pertaining to dialogues \cite{longchat2023} and multi-document question answering \cite{liu2023lost}, techniques like retrieval-augmented generation (RAG) are prevalent, and therefore explicitly isolating the length extrapolation issue poses a challenge.

To address these aforementioned problems, we present Lissard, a benchmark designed to evaluate the ability of models on tasks that require the use of repetitive simple rules, whose difficulty increases with respect to the sequence length. By incorporating varying degrees of difficulty within the same tasks, Lissard facilitates the identification of a models' breaking points. Given the syntactic nature of the datasets, researchers have the capability to generate new examples and increase the task difficulty, thus making it more challenging for newer and more capable models to be evaluated effectively.
This flexibility also mitigates the contamination problem -- where models may inadvertently be exposed to test datasets during their training \cite{ahuja2023mega,li2023task} -- since synthetic datasets can be generated as needed, a advantage over traditional, manually curated datasets.

Our analysis, which includes evaluations on proprietary models such as GPT-3.5-Turbo and GPT-4~\cite{openai2023gpt4}, as well as open-source ones like Mistral~\cite{jiang2023mistral}, reveals a common trend among them. As illustrated in Figure~\ref{fig:gpt_4_all_tasks}, our findings underscore that irrespective of their architectures and parameter counts, all examined models demonstrate a performance degradation with increasing length, controlled by the number of key entities (see their definition in Table~\ref{tab:key_entity_bins}), required to solve the tasks. This indicates a common point of failure in generalization for LLMs, even for sequence lengths that are considerably shorter in terms of tokens than those seen during their pretraining or fine-tuning phases.


\section{Related Work}


The challenge of length extrapolation in the domain of natural language processing has been a persistent and long-standing issue. An array of studies has demonstrated that neural architectures encounter difficulties when confronted with sequences of longer than those they encountered during their training \cite{lake2018generalization,livska2018memorize,keysers2019measuring,dubois-etal-2020-location, nogueira2021investigating, welleck2022symbolic,lewkowycz2022solving, delétang2023neural,zhou2023algorithms}. Despite efforts to expand the context window in LLMs, this issue persists, particularly when tackling tasks involving complex reasoning~\cite{anil2022exploring}. 


Recent endeavors have been undertaken to enhance the  general performance of LLMs by employing prompt engineering techniques and by developing novel decoding methods aimed at expanding their capacity to extrapolate effectively over lengthy sequences of tokens.
For instance, Nie et al.\cite{nye2021work} introduced the concept of a "scratchpad" that enables the model to generate draft responses in natural language before producing the final output. To assess the performance of this method, a range of tasks were employed, including math and coding tasks.
Moreover, studies by Wei et al. \cite{ wei2023chainofthought} and Zhou et al.~\cite{zhou2023leasttomost} demonstrated improvements by configuring the model to generate explanations for problem-solving and breaking down tasks into multiple interactive steps. These enhancements were particularly noticeable in tasks requiring the ability to extrapolate, such as last-letter concatenation (symbolic manipulation), SCAN~\cite{lake2018generalization} (compositional generalization), and mathematical reasoning.
Additionally, Bueno et al.~\cite{bueno2022induced} showed that utilizing markups tokens as position representations help the model to generalize to longer sequences in tasks related to mathematical addition and compositional generalization. Han et al.~\cite{han2023lminfinite} devised a decoding method to improve generalization over extended sequences.


In addition to techniques for customizing prompts, recent research has explored modifying the position encoding function of the original transformer architecture to enhance its extrapolation capabilities \cite{press2022train, chi2022kerple,chi-etal-2023-dissecting, li2023functional,qin2023exploring,chen2023extending}. For instance, Kazemnejad et al.~\cite{kazemnejad2023impact} conducted an evaluation of commonly used positional encoding methods, finding that omitting positional encoding altogether yielded superior results.


The studies above provide evidence of multiple approaches that have been developed to address the challenge of extrapolation. Nonetheless, there is a notable absence of research focusing on development of diverse and standardized datasets that assess the generation and synthesis of extended text sequences produced by neural models. This gap is particularly significant when considering that many of the classical datasets available in question may have already been used into the training of large language models.


\section{Datasets Description}

Our benchmark draws from both existing tasks from BIG-bench~\cite{srivastava2023beyond} and newly created ones. We decided not to include classic datasets (e.g., SCAN) in the analysis, as its test set present in a public repository and many of its solutions detailed in scientific publications may have already been seen by LLMs.

We categorized the datasets into two groups: ``Input Extrapolation'' and ``Generation Extrapolation''. In the first category, tasks demand skills in information extraction, simple arithmetic, memorization, and logical deduction. The goal is not for the model to generate long sequences of text as output, but rather for it to process lengthy chains of information to deduce intrinsic rules.

On the other hand, the category ``Generation Extrapolation'' encompasses tasks that, despite also requiring memorization, logic and deduction, require the model needs to generate long sequences of tokens as output.
Tasks in this category expose the limits of models in terms of information processing that normally comprises a short input and the generation of long sequences of tokens based on logical rules.

In Tables \ref{tab:summary_tasks} and \ref{tab:key_entity_bins} we summarize of the tasks in Lissard. In Table \ref{tab:summary_tasks}, we show examples of input and output, while in Table \ref{tab:key_entity_bins} we describe the key entities that control the complexity and the respective lengths in each bin of Figure~\ref{fig:gpt_4_all_tasks}. For the in-context few-shot examples used during model evaluation, we selected samples contained in the first bin, as these contain the smallest lengths.

The following sections describe the tasks and how evaluation was performed.

\begin{table}
\centering
\begin{tabular}{p{0.25\linewidth}p{0.40\linewidth}p{0.15\linewidth}}
\toprule
\textbf{Task} & \textbf{Input Example}& \textbf{Output Example} \\
\midrule    
Object Counting & I have a flute, and a drum & 2 \\
\midrule    
Logic Grid  & here are 2 houses next to each other, numbered 1 on the left and 2 on the right. There is one person living in each house (...) & 2 \\
\midrule    
Dyck Language  & [[] & ] \\
\midrule        
List Intersection & A: densil, rodriquez, vonetta / B: rodriquez, clarabelle, densil & rodriquez \\
\midrule
Last-Letter-Concat & Mahir Yolisma Saish  & r a h \\
\midrule    
Repeat Copy Logic  & A watermelon has 2 seeds. Repeat they're delicious once for every seed & they're delicious they're delicious \\
\midrule    
Vowel grouping & E A O E & EA,AO,OE \\
\bottomrule
\end{tabular}
    \caption{Summary of the tasks in the Lissard benchmark.}
    \label{tab:summary_tasks} 
\end{table}

\begin{table}
\centering
\begin{tabular}{p{0.2\linewidth}p{0.25\linewidth}p{0.1\linewidth}p{0.1\linewidth}p{0.1\linewidth}p{0.1\linewidth}}
    \toprule
    \textbf{Task} & \textbf{Key Entity}& \textbf{Bin 1}& \textbf{Bin 2}&\textbf{Bin 3}&\textbf{Bin 4}\\
    \midrule
    
    Last-Letter-Concat & Names & 1-8 & 8-15 & 15-22 & 22-30 \\
    \midrule
    Dyck Language & Symbols & 1-26 & 26-50 &50-74&74-99 \\
    \midrule
    Logic Grid & Houses & 1-2&2-3&3-4&4-5 \\
    \midrule
    Repeat Copy Logic & Number of repetitions & 1-9&9-17&17-25&25-33 \\
    \midrule
    Object Counting & Objects & 1-7&7-12&12-17&17-23 \\
    \midrule
    List Intersection & Items in lists A and B & 1-24&24-46&46-68&68-91 \\
    \midrule
    Vowel grouping & Number of letters in the sequence  & 1-23&23-45&45-67&67-89 \\
    \bottomrule
    \end{tabular}
    \caption{Key entities of each task and their respective ranges in each bin in Figure~\ref{fig:gpt_4_all_tasks}a.}
    \label{tab:key_entity_bins} 
\end{table}

\subsection{Input Extrapolation}
\subsubsection{Object Counting}
The main goal of this task is to assess the proficiency in object counting within sequences, as shown in Table~\ref{tab:object_count}. The input to the model is a sequence comprising a list of objects paired with their respective quantities and the expected output is a string with the total count of objects. Diverging from the original BIG-bench task that exclusively encompasses the enumeration of objects from predetermined categories like fruits, vegetables, or musical instruments, our method comprises object counting across different categories. 


\begin{table}
\centering
\begin{tabular}{p{0.8\linewidth}C{0.13\linewidth}}
    \toprule
    \textbf{Input} & \textbf{Target} \\
    \midrule    
    I have three onions, two potatoes, and a cabbage. & 6 \\
    \midrule
    
    I have a couch, two microwaves, an oven, a toaster, a fridge, a table, two beds, and a car. & 10 \\
    \bottomrule
    \end{tabular}
    \caption{Examples of input and target sequences of the Object Counting task.}
    \label{tab:object_count} 
\end{table}

\subsubsection{Logic Grid}
This task aims to evaluate the proficiency of language models in deduction, logic, information extraction, and memory. The dataset include residences, each occupied by an individual with distinct characteristics. Subsequently, spatial correlations between characters and their attributes are identified. The goal is for the model to determine the residence of a person with specific traits. Table~\ref{tab:logic_grid} illustrates an example from the dataset. 


\begin{table}
\centering
\begin{tabular}{p{0.8\linewidth}C{0.12\linewidth}}
    \toprule
    \textbf{Input} & \textbf{Target} \\
    \midrule    
   There are 2 houses next to each other, numbered 1 on the left and 2 on the right. There is one person living in each house. The people in these houses have different characteristics:\newline
    - Each person plays a different musical instrument: one is a violinist and one is a guitarist\newline
    - Each person is eating a different kind of fruit: one is eating oranges and one is eating apples\newline
     - Everyone likes a different kind of book: one is a romance book lover and one is a science fiction book fanatic\newline
    - Each person has a favorite drink: one likes milk and one is a root beer lover\newline

    Clue(s):\newline
    1. The person who is eating apples lives in the second house.
    2. The romance book lover lives in the second house.
    3. The violinist lives somewhere to the left of the person who likes milk.

What is the number of the house where the person who likes milk lives? & 2 \\
    \bottomrule
    \end{tabular}
    \caption{An example of input and target sequences of the Logic Grid task.}
    \label{tab:logic_grid} 
\end{table}

\subsubsection{Dyck Language}
Dyck-n functions as a representation of languages characterized by hierarchical structures. Here we use Dyck languages to measure the capacity of models to capture hierarchical representations. 
Therefore, the goal of this task involves predicting the order of closing parentheses within a Dyck-n word, given a unfinished sequence of parethesis. This task serves as an investigation into the hierarchical nature of language and the capabilities of computational models in this domain.

\begin{table}
\centering
\begin{tabular}{p{0.3\linewidth}C{0.13\linewidth}}
    \toprule
    \textbf{Input} & \textbf{Target} \\
    \midrule    
     $[ < < >$ & $> ]$  \\
    \midrule
     $( < < < ( < < > > ) > > >$ & $)$  \\
    \bottomrule
    \end{tabular}
    \caption{Examples of input and target sequences of the Dyck Language task.}
    \label{tab:dyck_language} 
\end{table}

\subsection{Generation Extrapolation}

\subsubsection{List Intersection}

In this task the objective is to evaluate the models' ability to track and identify intersections between two given lists.
The items in the lists are from names of people randomly sampled from Name Census.\footnote{https://namecensus.com/} The lists have equal sizes but the number of overlapping items varies. The target output is the names in common, sorted alphabetically. If there are no items in common, "None" must be returned. 

\begin{table}
\centering
\begin{tabular}{p{0.4\linewidth}p{0.3\linewidth}}
    \toprule
    \textbf{Input} & \textbf{Target} \\
    \midrule    
     A: densil rodriquez vonetta  \\ B: clarabelle lareyna maeli & None \\ 
    \midrule
    A: densil rodriquez vonetta \\ B: densil rodriquez vonetta  &  densil rodriquez vonetta\\
    \midrule
    A: densil rodriquez vonetta \\ B: rodriquez clarabelle densil  &  densil rodriquez\\
    \midrule
    A: densil rodriquez vonetta \\ B: rodriquez clarabelle maeli  &  rodriquez\\
     \bottomrule
     \end{tabular}
    \caption{Examples of input and target sequences of the List Intersection task.}
    \label{tab:list_intersection} 
\end{table}

\subsubsection{Last Letter Concatenation}
The Last Letter Concatenation task, as formulated in the Chain-of-Thought work \cite{wei2023chainofthought}, involves concatenating the last letter of each word within an input sequence comprised of random names. Table~\ref{tab:last_letter_concat} provides an illustrative instance of the dataset, where the input sequence comprises randomly generated names obtained through Name Census. 

In constructing our dataset, we applied a comparable methodology but expanded the sample length to encompass sequences with an increase of up to thirty names.

\begin{table}
\centering
\begin{tabular}{p{0.65\linewidth}p{0.25\linewidth}}
    \toprule
    \textbf{Input} & \textbf{Target} \\
    \midrule    
     Halona Grainne Lodena Annalecia & a e a a \\
    \midrule
     Arcenio Nashad Alishaba Mishonda Dagmawi Aidin Belmeda Eloida Olwyn Devente Takeria Raiyne Marguret Jabarri Ariam Nolen Sacha Kameela Remedy Suresh  & o d a a i n a a n e a e t i m n a a y h \\

    \bottomrule
    \end{tabular}
    \caption{Examples of input and target sequences of the Last Letter Concatenation task.}
    \label{tab:last_letter_concat} 
\end{table}

\subsubsection{Repeat Copy Logic}
The task proposed by the BIG-bench benchmark assesses models' competence in comprehending and executing instructions encompassing repetitions, text-to-copy, basic logic, and conditionals.
The dataset incorporates a repetition factor, which is interesting for evaluating extrapolation capacity. For example, given the input \textbf{``Repeat 5 times hello world''}, the expected output should be \textbf{``hello world hello world hello world hello world hello world''}.

Our methodology for evaluating extrapolation encompassed the following steps:
\begin{itemize}[label={}]
    \item i) We obtained responses to all input sequences made available in the BIG-bench repository\footnote{\url{https://github.com/google/BIG-bench/tree/main/bigbench/benchmark_tasks/repeat_copy_logic}}, without alterations.
    \item ii) We collected the responses and retained only those instructions correctly answered by GPT-4.This methodology was employed due to instances where the model fails to generate the correct answer even when the number of key entities is small. For instance, out of the original 32 questions, only 17 were answered correctly by GPT-4. Through this approach, we can pinpoint the breaking point by increasing the number of repetitions in each question.
    \item iii) For each selected instruction, we conducted extrapolations of the repetition factor, ranging from 1 to up to 33.
\end{itemize}

During step ii) Comparison by exatch match was adopted, in addition, from these 17 questions, we randomly selected 15. In step iii), we utilized integers as the repetition factor.

\subsubsection{Vowel Grouping}
The task at hand aims to assess the capacity of models in generating overlapping element windows. Provided with a list of uppercase vowels separated by whitespace, the language model is expected to produce groups of 2 elements with overlaps (n - k + 1). These groups must maintain the original order of the letters as given in the sequence, akin to the generation of context windows. Hence, this task necessitates the models to possess the ability to memorize and systematically monitor sequences. Table~\ref{tab:example_vowel} shows examples of the task.

\begin{table}
\centering
\begin{tabular}{p{0.4\linewidth}C{0.55\linewidth}}
    \toprule
    \textbf{Input} & \textbf{Target}  \\
    \midrule    
  I I E & II,IE \\
    \midrule
  E E E O E U U U & EE,EE,EO,OE,EU,UU,UU \\
    \midrule
  E U I E E E O E E A A I A &  EU,UI,IE,EE,EE,EO,OE,EE,EA,AA,AI,IA \\
    \bottomrule
    \end{tabular}
    \caption{Examples of input and target sequences of the Vowel Grouping task.}
    \label{tab:example_vowel} 
\end{table}

\section{Baseline Methods}

The evaluation of each task involved analyzing responses from gpt-3.5-turbo-1106, gpt4-0613, mistral-7b-v0.2, and mixtral-8x7b-v0.1. We used greedy decoding for all tasks and did not observe repetition issues. We used a predefined prompt template for each task with the in-context learning approach, i.e. for the tasks ``Object counting'', ``Dyck language'', ``Find intersection'', ``Last Letter Concat'' and ``Vowel Grouping'' we provide four in-context examples to the model, while for the ``Repeat Copy Logic'' and ``Logic Grid'' tasks we use a single in-context example, as these last two tasks already contain descriptive instructions we saw no need to add more examples to the prompt. After executing the tasks, we measured exact match performance.

We did not experimented with more sophisticated prompting strategies, such as Chain-of-Thought or special markup tokens, as our goal is to establish simple baselines. Due to the costs of using paid APIs, we limited our tests to around 300 examples for each task, and for each bin (see Table~\ref{tab:key_entity_bins}) we randomly selected 75 examples to ensure a fair evaluation between the length partitions. 

\section{Results}

\begin{figure}[h] 
    \begin{center}
        \includegraphics[width=1\textwidth]{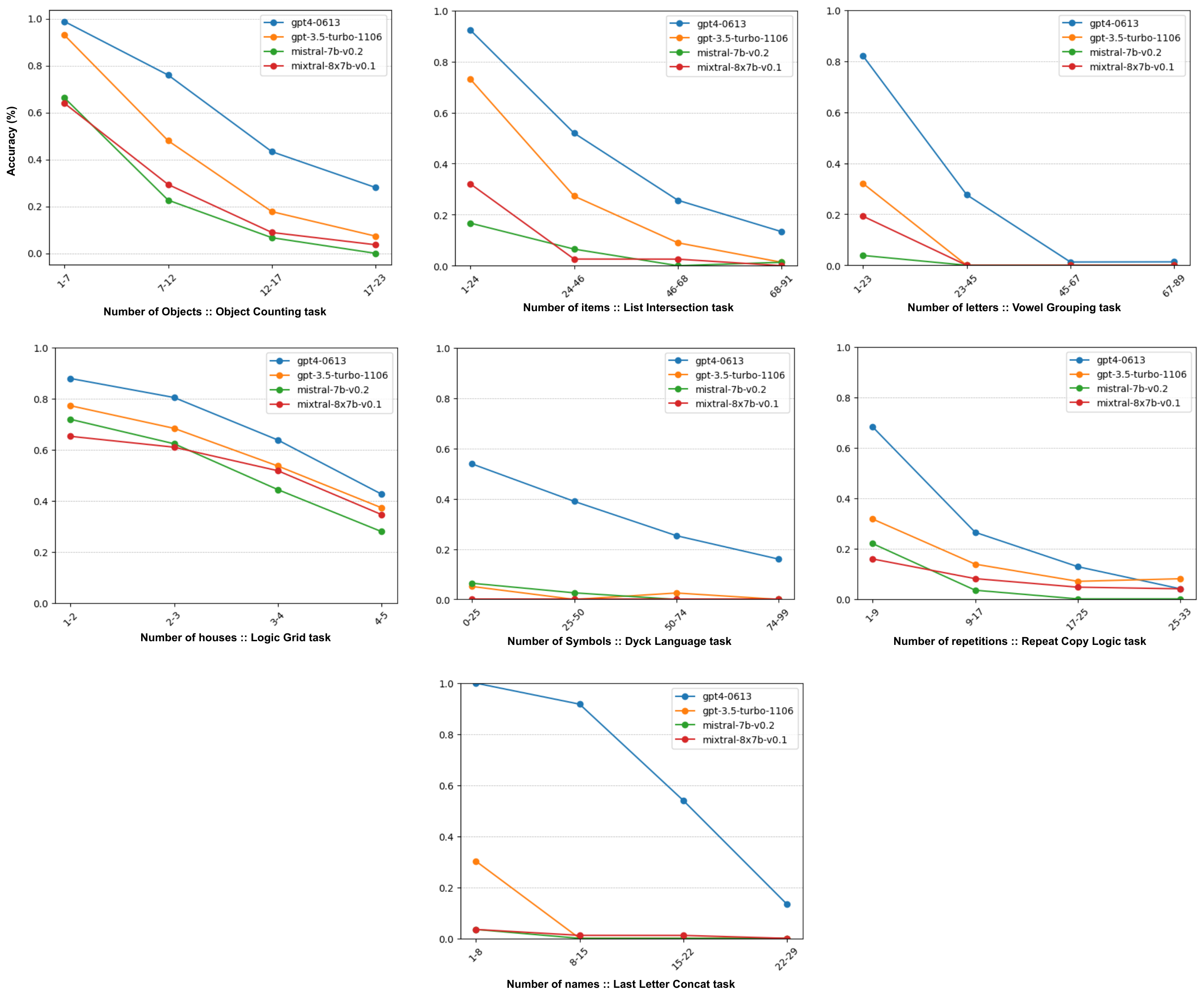} 
    \end{center}
    \caption{Results per task.}
    \label{fig:all_results}
\end{figure}

Results for all tasks are shown in Figure~\ref{fig:all_results}.
Overall, there is a gradual performance decline in language models across all tasks as complexity, measured by the number of key entities in the input sequence, increases. For instance, in the ``Object Counting'' task, when presented with inputs containing 1 to 7 objects, GPT-4 and GPT-3.5 models achieve approximately 100\% accuracy. However, their accuracy drops below 50\% when confronted with sequences with 12 to 17 objects. Analogous trends in performance were identified in the ``Last Letter Concat'' and ``List Intersection'' tasks.

However, in the context of the ``Repeat Copy Logic'' task, a pronounced decrease in performance was observed when extrapolating from 1 to 9 repetitions to 9-17 repetitions. Specifically, while achieving a 70\% accuracy rate for the former range, the latter range exhibited a performance decline to approximately 30\%. Conversely, within the ``Logic Grid'' task, an inverse pattern emerged, characterized by an average decrease of 10\% in performance across iterations.

Across all tasks, we observed that the open-source models from Mistral consistently lag behind GPT-4 in performance but demonstrated competitive capabilities compared to the GPT-3.5, particularly evident in tasks such as ``Dyck language'', ``Logic Grid'' and ``Repeat Copy Logic''. However both the open-source models and GPT-3.5 have a poor performance across most tasks when the number of key entities become longer. Despite Mixtral-8x-7b having approximately six times more parameters than Mistral-7B, their performances are comparable across tasks.

\noindent\textbf{Input Extrapolation vs Generation Extrapolation:}
Despite the general trend of models experiencing a decrease in performance as the number of the key entity increases, we observe that tasks falling under the ``Generation Extrapolation'' category are more challenging for language models. For example, within the ``Input Extrapolation'' category, GPT-4 achieves an accuracy below 20\% in the final bin exclusively for the "Dyck Language" task. However, in the ``Generation Extrapolation'' category, such low levels of accuracy are observed across all four tasks.



\section{Conclusion}

We presented a benchmark to evaluate the ability of language models to deal with long texts. Our approach distinguishes itself from existing benchmarks through the introduction of a control mechanism, which we refer to as "key entities." This mechanism enables us to systematically increase task complexity in tandem with sequence length. Furthermore, the ability to solve these tasks is predicated on the repeated application of simple rules, providing more control and enabling a detailed analysis of model performance in relation to the frequency of rule application. This contrasts with benchmarks that rely on lengthy natural language texts, where the relationship between text length and task difficulty may become obscured. Despite the apparent simplicity of these tasks, they reveal significant limitations in state-of-the-art LLMs concerning the processing and generation of text as lengths increase. We have made both the dataset and the scripts for generating it publicly available, hoping to estimulate the research community to improve language model performance with respect to these identified limitations.


\bibliographystyle{abbrv}
\bibliography{main}

\end{document}